\documentclass[10pt,twocolumn,letterpaper]{article}

\usepackage{cvpr}
\usepackage{times}
\usepackage{epsfig}
\usepackage{graphicx}
\usepackage{amsmath}
\usepackage{amssymb}
\usepackage{bbm}
\usepackage{caption}

\usepackage[pagebackref=true,breaklinks=true,letterpaper=true,colorlinks,bookmarks=false]{hyperref}
\cvprfinalcopy % *** Uncomment this line for the final submission

\def\cvprPaperID{2154} % *** Enter the CVPR Paper ID here

\ifcvprfinal\fi
\begin{document}
%%%%%%%%% TITLE
\title{\vspace*{-1.0 cm} {``Double-DIP''} :\\
% \title{{``Double-DIP''} :\\
Unsupervised Image Decomposition \ \  via \ \ {Coupled Deep-Image-Priors}}
\author{Yossi Gandelsman \qquad Assaf Shocher \qquad Michal Irani\\
Dept. of Computer Science and Applied Mathematics\\
The Weizmann Institute of Science, Israel\\
\textit{\textbf{Project Website}: \href{http://www.wisdom.weizmann.ac.il/~vision/DoubleDIP/index.html}{www.wisdom.weizmann.ac.il/$\sim$vision/DoubleDIP/}}
% {\tt\small \{yosef.gandelsman, assaf.shocher, michal.irani\}@weizmann.ac.il}
}
\twocolumn[{%
\renewcommand\twocolumn[1][]{#1}%
\maketitle
\vspace*{-0.7cm}
\begin{center}
\centering
\vspace*{-0.2cm}
\includegraphics[width=0.95\textwidth]{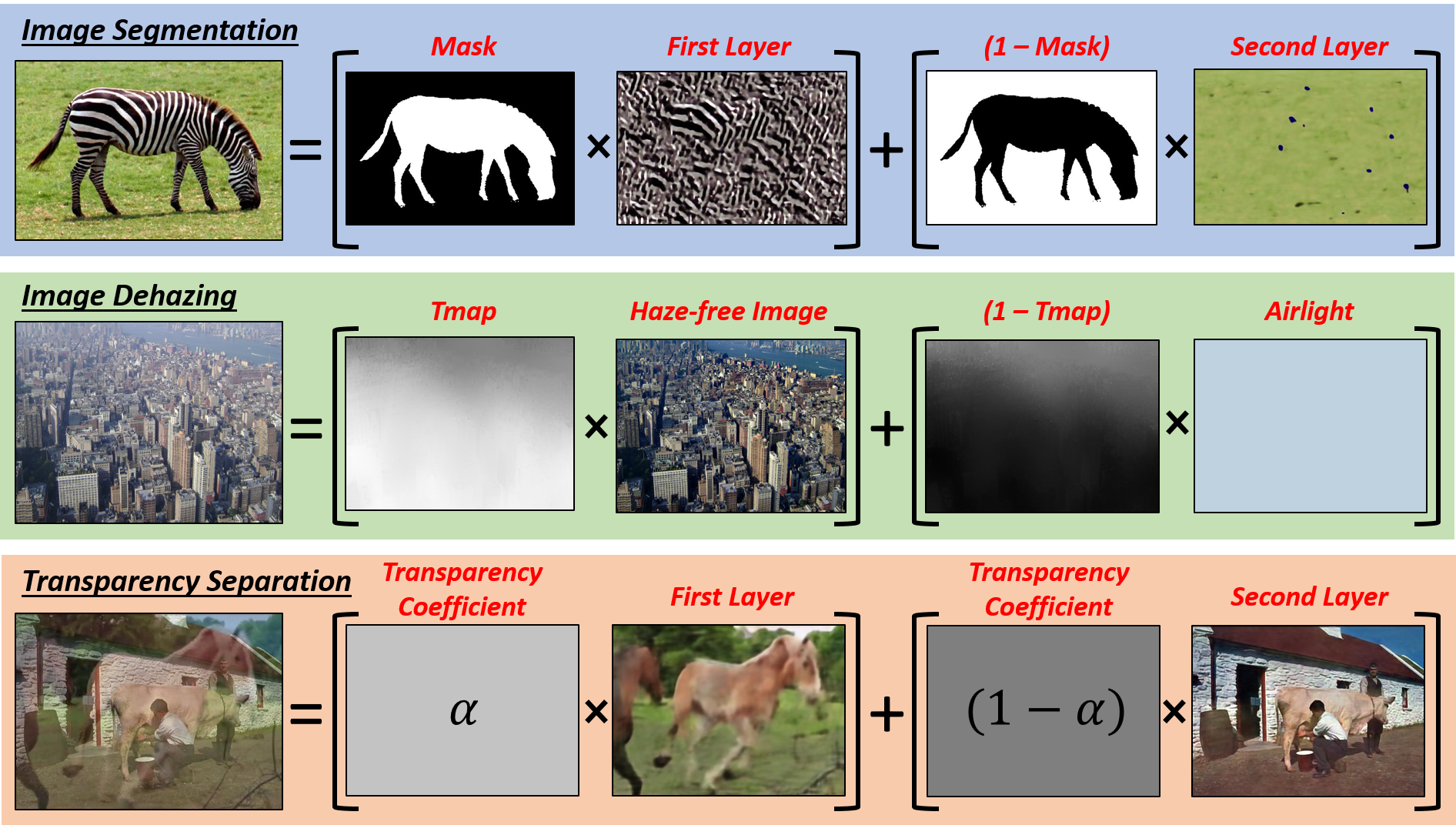}
%***************************************************\\
%Suggestion: adding here watermark removal example.\\
%***************************************************\\
\captionof{figure}{\textbf{A unified framework for image decomposition.} An image can be viewed as a mixture of ``simpler'' layers. Decomposing an image into such layers provides a unified framework for many seemingly unrelated vision tasks (e.g., segmentation, dehazing, transparency separation). Such a decomposition can be achieved using ``Double-DIP''.}
%{Image decomposition via DoubleDIP.} Various computer-vision tasks can be presented as a decomposition challenge, of an image to distinct layers. DoubleDIP exploits this fact and offers a unified approach of "solving by separating".}
\label{fig:decomposition}
\vspace*{0.1cm}
\end{center}%
}]
%\thispagestyle{empty}
%%%%%%%%% ABSTRACT
\begin{abstract}
\vspace*{-0.2cm}
%An image can be viewed as a mixture of multiple layers. 
Many seemingly unrelated computer vision tasks can be viewed as a special case of image decomposition into separate layers.  For example, image segmentation (separation into foreground and background layers); transparent layer separation (into reflection and transmission layers); Image dehazing (separation into a clear image and a haze map), and more. In this paper we propose a unified framework for unsupervised layer decomposition of a single image, based on coupled ``Deep-image-Prior'' (DIP) networks. 
It was shown~\cite{UlyanovVL17} that the structure of a single DIP generator network is sufficient to capture the low-level statistics of a single image. We show that coupling multiple such DIPs provides a powerful tool for decomposing images into their basic components, for a wide variety of applications. This capability stems from the fact that the internal statistics of a mixture of layers is more complex than the statistics of each of its individual components. We show the power of this approach for Image-Dehazing, Fg/Bg Segmentation, Watermark-Removal, Transparency Separation in images and video, and more. These capabilities are achieved in a totally unsupervised way, with no training examples other than the input image/video itself.
\footnote{Code will be made publicly available.}
%\newpage
\end{abstract}
\vspace{-0.5 cm}

\begin{figure}
\vspace*{-0.4cm}
  \centering
  \hspace*{-0.2cm} \includegraphics[width=\columnwidth]{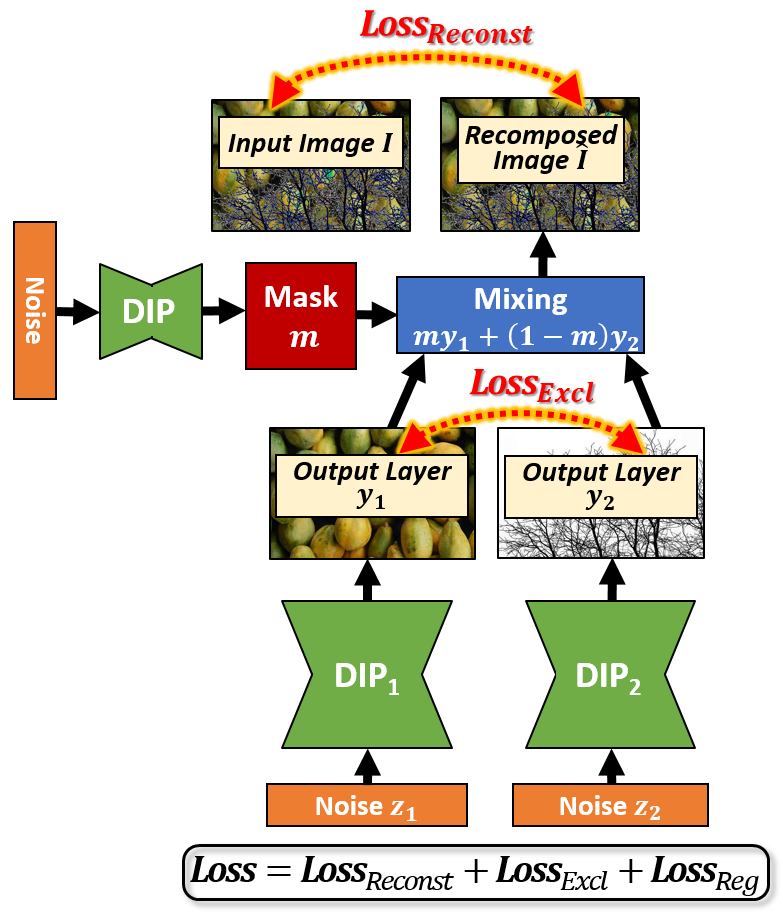}
\caption{\label{fig:DoubleDip}
%\textbf{Image-specific information -- often predicted only internally.}
\textbf{Double-DIP Framework.} 
\emph{Two Deep-Image-Prior networks (DIP$_1$ $\&$ DIP$_2$)  jointly decompose an input image $I$ into its layers ($y_1$ $\&$ $y_2$). Mixing those layers back according to a learned mask $m$,  reconstructs an image $\hat{I}$$\approx$$I$.}
%\emph{Two or more deep-image-prior networks are applied to noise to reconstruct given image.  Each network produces one layer of the decomposed output. Deep image priors, together with output coherent decomposition of the original image.}
}
\vspace{-0.5cm}
\end{figure}

%%%%%%%%% BODY TEXT
\begin{figure*}
\vspace*{-0.2cm}
  \centering
  \hspace*{-0.2cm} \includegraphics[width=1.0\textwidth]{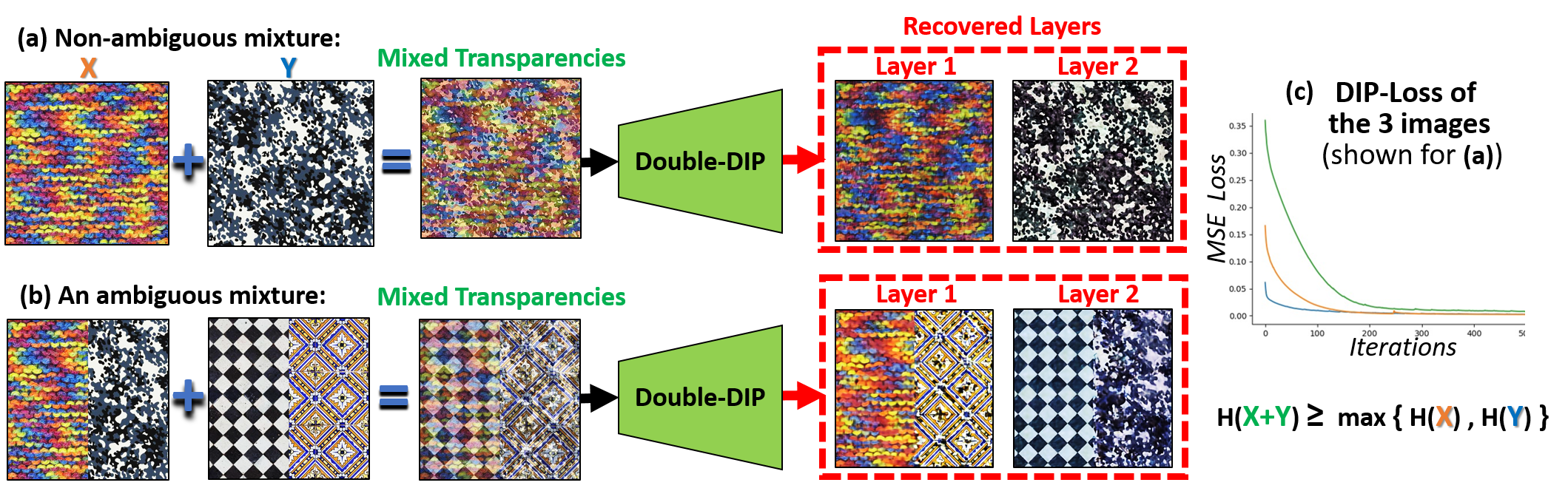}
\caption{\label{fig:textureSumLoss}
\textbf{The complexity of mixtures of layers vs. the simplicity of the individual components.}
\emph{(See text for  explanation).}
}
\vspace{-0.2cm}
\end{figure*}

\section{Introduction}
\label{sec:introduction}
%An image can be viewed as a mixture of multiple components.
Various computer vision tasks aim to decompose an image into its individual components. In image/video segmentation, the task is to decompose the image into meaningful sub-regions, such as foreground and background~\cite{AlpertGBB07,eccv08_bagon_boiman_irani,Faktor:ICCV'2013,KIM2005172, Rother:2004:GIF:1015706.1015720}. In transparency separation, the task is to separate the image into its superimposed reflection and transmission~\cite{transparencyAvidan:CVPR2000,Sarel,transparencyLevin:2007,Dekel17}. Such transparency can be a result of accidental physical reflections, or due to intentional transparent overlays (e.g., watermarks). In image dehazing~\cite{Bahat16,He2011,Fattal2014,Ren-ECCV-2016,NonLocalImageDehazing}, the goal is to separate a hazy/foggy image into its underlying haze-free image and the obscuring haze/fog layers (airlight and transmission map). Fig.~\ref{fig:decomposition} shows how all these very different tasks can be casted into a single unified framework of \emph{layer-decomposition}. What is common to all these decompositions is the fact that the \emph{distribution of small patches} within each separate layer is ``simpler'' (more uniform) than in the original mixed image, resulting in \emph{strong internal self-similarity}.
%(see Zebra image).
%In other words, there is \emph{strong internal self-similarity} of patches within each layer.

Small image patches (e.g., 5x5, 7x7)
%(e.g., $5 \times 5$, $7 \times 7$)
have been shown to repeat abundantly inside a single natural image~\cite{Glasner:ICCV09,Zontak2011InternalSO}. This strong internal patch recurrence was exploited for solving a large variety of computer vision  tasks~\cite{NLM2005,Dabov2007,Elad2006,efros1999texture,Glasner:ICCV09,Simakov2008,Pritch:ICCV09,Barnes09,Cho_TPAMI2010}.
%, including images denoising~\cite{NLM2005,Dabov2007,Elad2006}, texture synthesis~\cite{efros1999texture}, image completion and retargeting~\cite{Simakov2008,Pritch:ICCV09,Barnes09,Cho_TPAMI2010}, super-resolution~\cite{Glasner:ICCV09,SelfExSR:CVPR2015}, and more.
It was also shown that the empirical entropy of patches inside a \emph{single image} is much smaller than the  entropy in a \emph{collection of  images}~\cite{Zontak2011InternalSO}.
It was further observed by~\cite{eccv08_bagon_boiman_irani} that 
%different segments within the same image tend to have different distributions of small image regions. They showed that 
the \emph{empirical entropy} of small image regions composing a segment, is smaller than the {\em empirical
cross-entropy} of regions across different segments within the same image.
This observation has been successfully used for \emph{unsupervised} image segmentation~\cite{eccv08_bagon_boiman_irani,Faktor:ICCV'2013}.
Finally, it was observed~\cite{Bahat16} that 
%the strong internal similarity of patches within an image significantly diminishes when the image is taken under poor visibility conditions (e.g., haze, fog, underwater). 
the distribution of patches in a \emph{hazy image} tends to be more diverse (weaker internal patch similarity) than in its underlying \emph{haze-free} image. This observation was exploited by~\cite{Bahat16} for blind image dehazing.

In this paper we combine the power of the internal patch recurrence (its strength in solving unsupervised tasks), with the power of Deep-Learning. We propose an \emph{unsupervised} Deep framework for decomposing a single image into its layers, such that the distribution of ``image elements'' within each layer is ``simple''.
We build on top of the ``Deep Image Prior'' (DIP) work of Ulyanov \etal~\cite{UlyanovVL17}.
They showed that the structure of a \emph{single} DIP generator network is sufficient to capture the low-level statistics of a \emph{single} natural image.
The input to the DIP network is \emph{random noise}, and it trains to \emph{reconstruct a single image} (which serves as its sole output training example).
%The DIP network training is unsupervised, is initialized by \emph{random noise} and uses the input image as its sole output training example.
This network was shown to be quite powerful for solving inverse problems like denoising, super-resolution and inpainting, in an unsupervised way.
%
%
%In fact, we observe that \emph{similar} small patches inside the same image tend to be all generated by a single DIP network. In other words, when employing multiple DIPs, each DIP tends to capture different components of the internal statistics of the input image. This can be explained by the fact that a single DIP is a \emph{fully convolutional} networks, hence its filter weights are shared across the entire spatial extent of the image. This promotes self-similarity of patches in the output of each DIP.
%
%We refer to such a consorted effort of two or more DIPs as  \emph(``Double-DIP'') (``Coupled-DIPs'' when more than two). The separation capability of Double-DIP stems from the fact that the internal statistics of a mixture of layers (the input image) is more complex than the statistics of each of its individual components.

We observe that when employing \emph{a combination of multiple DIPs} to reconstruct an image, those DIPs tend to ``split'' the  image, such that the patch distribution of each DIP output is ``simple''.
Our approach for \emph{unsupervised multi-task layer decomposition}  is thus based on
%a consorted effort of two or more DIPs,
{a combination of multiple (two or more) DIPs}
which we coin ``Double-DIP''. 
We demonstrate the applicability of this approach to a wide range of computer vision tasks, including Image-Dehazing, Fg/Bg Segmentation of images and videos, Watermark Removal, and Transparency Separation in images and videos.
%The variations required in the proposed framework to achieve the different tasks are very minute (explained in later sections).
%This observation allows as to solve wide range of vision tasks without external supervision and with only small variations in the training optimization processes.

Double-DIP is \emph{general-purpose} and caters many different applications.
\emph{Special-purpose} methods designed for one specific task may outperform Double-DIP on their own challenge. However, to the best of our knowledge, this is the first framework that is able to handle well such a large variety of image-decomposition tasks.
%(and in an unsupervised way).
Moreover, in some tasks (e.g., image dehazing), Double-DIP  achieves comparable and even better results than leading methods.

\section{Overview of the Approach}
\label{sec:intuition}
%In this section we provide a high-level overview of our  approach, and try to convey its underlying intuition.
Observe the illustrative example in Fig.~\ref{fig:textureSumLoss}a. Two different textures, $X$ and $Y$, are mixed to form a more complex image $Z$ which exhibits layer transparency. The distribution of small patches and colors inside each pure texture is \emph{simpler} than the distribution of patches and colors in the combined image. Moreover, the similarity of patches across the two textures is very weak.
It is well known~\cite{CoverBook:2006} that if $X$ and $Y$ are two \emph{independent} random variables,  the entropy of their sum $Z=X+Y$ is larger than their individual entropies: $max\{H(X), H(Y )\} \leq H(Z)$. We leverage this fact to separate the image into its natural ``simpler'' components.

% We build on top of the work of
% Ulyanov \etal~\cite{UlyanovVL17}, who showed that the structure of a deep convolutional neural network captures the low-level statistics of a \emph{single} natural image. Their network is initialized by \emph{random noise} and uses the single image as its sole output for training. The learned weights constitute a deep image prior for that specific image.
% This was shown to be useful for inverse problems like denoising, super-resolution, and inpainting.

\subsection{Single DIP vs. Coupled DIPs}

Let's see what happens when a DIP network is used to learn \emph{pure images} versus \emph{mixed images}. The graph in Fig.~\ref{fig:textureSumLoss}.c shows the MSE Reconstruction Loss of a \emph{single} DIP network, as a function of time (training iterations), for each of the 3 images in  Fig.~\ref{fig:textureSumLoss}.a: (i)~the orange plot is the loss of a DIP  trained to reconstruct the texture image X, (ii)~the blue plot -- a DIP trained to reconstruct the texture Y, and (iii)~the green plot -- a DIP  trained to reconstruct their superimposed mixture (image transparency). Note the larger loss and longer convergence time of the mixed image, compared to the loss of its individual components. In fact, the loss of the mixed image is larger than the \emph{sum} of the two individual losses. We attribute this behavior to the fact that the distribution of patches in the mixed image is more complex and diverse (larger entropy; smaller internal self-similarity) than in any of its individual components. 
%This training behavior repeated itself also in the case of the non-overlapping multi-segment image (graph omitted).

While these are pure textures, the same behavior  holds also for mixtures of  natural images.
The \emph{internal self-similarity} of patches inside a single natural image tends to be much stronger than the patch similarity \emph{across different  images}~\cite{Zontak2011InternalSO}. 
%
%Similarly,    the \emph{empirical entropy} of small regions composing an image segment is smaller than their {\em empirical
%cross-entropy} across different segments in the same image~\cite{eccv08_bagon_boiman_irani}.
%
We repeated the above experiment for a large collection of \emph{natural images}: We \emph{randomly sampled} 100 pairs of images from the BSD100 dataset~\cite{BSD100}, and mixed each pair. For each image pair we trained a DIP to learn the mixed image and each of the individual images. The same behavior exhibited in the graph of Fig.~\ref{fig:textureSumLoss}.c repeated also in the case of natural images -- interestingly, with an even larger gap  between the loss of the mixed image and its individual components  (see graph in the \href{http://www.wisdom.weizmann.ac.il/~vision/DoubleDIP/index.html}{project website}).
%In fact, the \emph{gap} between the mixed image and its individual components was found to be even larger in the natural image experiment.

We performed a similar experiment for non-overlapping image segments. It was observed~\cite{eccv08_bagon_boiman_irani} that the \emph{empirical entropy} of small regions composing an image segment is smaller than their {\em empirical
cross-entropy} across different segments in the same image. 
We {randomly sampled} 100 pairs of images from the BSD100 dataset. For each pair we generated a new image, whose left side is the left side of one image, and whose right side is the right side of the second image. We  trained a DIP to learn the mixed image and each of the individual components. The graph behavior  of Fig.\ref{fig:textureSumLoss}.c repeated also in this case (see \href{http://www.wisdom.weizmann.ac.il/~vision/DoubleDIP/index.html}{project website}).

{We further observe that when \emph{multiple} DIPs train to \emph{jointly} reconstruct a single input image, they tend to ``split'' the  image patches among themselves.}
%, such that the patch distribution of each DIP output is as simple as possible.
%In fact, we observe that
Namely, \emph{similar} small patches inside the image tend to all be generated by a \emph{single} DIP network. In other words, each DIP captures different components of the internal statistics of the  image. We explain this behavior by the fact that a single DIP network is \emph{fully convolutional}, hence its filter weights are shared across the entire spatial extent of the image. This promotes self-similarity of patches in the output of each DIP.

The simplicity of the patch distribution in the output of a single DIP is further supported by the denoising experiments reported in~\cite{UlyanovVL17}. When a DIP was trained to reconstruct a \emph{noisy} image (high patch diversity/entropy), it was shown to generate along the way an \emph{intermediate clean} version of the image, before overfitting the noise.
%in later iterations.
 The clean image has higher internal patch similarity (smaller patch diversity/entropy), hence is simpler for the DIP to reconstruct.

Building on  these observations, we propose to decompose an image into its layers  
%using \emph{a combination of multiple (two or more) DIPs},
by \emph{combining multiple (two or more) DIPs},
which we call ``Double-DIP''.
Figs.~\ref{fig:textureSumLoss}.a,b show that when training 2 DIP networks to jointly recover the mixed texture transparency image (as the sum of their outputs), 
%of Fig.~\ref{fig:textureSumLoss}.a, 
each DIP outputs a coherent layer on its own. 
%(as the mixed sum of their outputs), each DIP outputs a coherent layer on its own. 
%In fact, they almost perfectly separate the mixed image into its original components (up to an additive  constant color ambiguity).

%although there are many possible decompositions of the transparency image into different components, the Double-DIP network each DIP network generates coherent and informative layer, and

%Fig.~\ref{fig:experiment} shows that the Double-DIP network
%manages to almost perfectly separate the mixed texture image into its original components (up to a global constant color ambiguity). 
%

\subsection{\mbox{Unified Multi-Task Decomposition Architecture}}
\label{sec:method}
%#####
\noindent
\textbf{What is a good image decomposition?} \ \
There are {infinitely} many possible decompositions of an image into layers. However, we suggest that a \emph{meaningful} decomposition
%be as simple as possible. In particular, it should
satisfies the following criteria: \
(i)~the recovered layers, when recombined, should yield the input image.
%This criterion can be enforced by minimizing a \emph{``Reconstruction Loss''}. \
% (using some pre-determined combination function).
(ii)~Each of the layers should be as ``simple'' as possible, namely, it should have a strong internal self-similarity of ``image elements''.
%This criterion can be obtained by using \emph{multiple DIPs} (one per layer). \
(iii)~The recovered layers should be as independent of each other (uncorrelated) as possible.
%This criterion can be enforced by introducing an \emph{``Exclusion Loss''} between the outputs of the different DIPs.

These criteria form the basis of our general-purpose Double-DIP architecture, illustrated in Fig.~\ref{fig:DoubleDip}.
The first criterion is enforced via a \emph{``Reconstruction Loss''}, which measures the error between the constructed image and the input image (see Fig.~\ref{fig:DoubleDip}). \
The second criterion is obtained by employing \emph{multiple DIPs} (one per layer). \
The third criterion is enforced by an \emph{``Exclusion Loss''} between the outputs of the different DIPs (minimizing their correlation).

Each DIP network (\textit{DIP}$_i$)
reconstructs a different layer $y_i$ of the input image $I$. The input to each \textit{DIP}$_i$ is randomly sampled uniform noise, $z_i$. The DIP outputs, $y_i$=\textit{DIP}$_i(z_i)$, are mixed using a weight mask $m$, to form a reconstructed image $\hat{I} = m \cdot y_1 + (1-m) \cdot y_2$, which should be as close as possible to the input image $I$.  
%The DIP outputs $y_i$ are the output of our algorithm.

%Some tasks require an additional weight mask to combine the different layers.
In some tasks the weight mask $m$ is simple and known, in other cases it needs to be learned (using an additional DIP). The learned mask $m$ may be uniform or spatially varying, continuous or binary.
These constraints on $m$ are task-dependant, and are enforced using a task-specific \emph{``Regularization Loss''}.
%The weight mask is also learned by a single DIP, but the constraints on the each weight mask is application-dependant, and is obtained by a ``Regularization Loss'' (e.g., in the segmentation task the weight mask has to be as close as possible to a binary image, while in the transparency task the T-map needs to be smooth).
%
%As shown in figure \ref{fig:DoubleDip}, the outputs of the networks $y_i = f_i(z_i)$ are the inputs to mixing function $m$, which is constructed of some combination of multiplication and addition of the layers.
%
%The optimization loss of our framework is therefore:
The optimization loss is therefore:
\vspace*{-0.1cm}
\begin{equation}
\label{eq:framwork_loss}
Loss = {Loss}_{Reconst} +  \alpha \cdot {Loss}_{Excl} + \beta \cdot {Loss}_{Reg}
\vspace*{-0.1cm}
\end{equation}
%Where $L_{Reconst}$ is \textit{reconstruction loss} norm function between the original image and the mixed output of the networks, $L_{Excl}$ is \textit{exclusion loss}, and $L_{Reg}$ is regularization term.
where ${Loss}_{Reconst}= \lVert I - \hat{I} \rVert$, and
${Loss}_{Excl}$  (the Exclusion loss) minimizes the correlation between the \emph{gradients} of $y_1$ and $y_2$ (as defined in~\cite{Zhang2018SingleIR}).
%The exclusion loss ${Loss}_{Excl}$ minimizes the correlation between $y_1$ and $y_2$. We use the images-exclusion loss presented in~\cite{Zhang2018SingleIR}, which minimizes cross-scaled gradient correlation between the outputs.
${Loss}_{Reg}$ is a task-specific mask regularization (e.g., in the segmentation task the mask $m$ has to be as close as possible to a binary image, while in the dehazing task the $t$-map is continuous and smooth).
We further apply guided filtering~\cite{He2010} on the learned mask $m$ to obtain a refined mask.\\

%
%The architecture used for each DIP is similar to the one used in \cite{UlyanovVL17}, with minor changes for some of the applications. In general, it is an hour-glass architecture with skip connections, similarly to \cite{RFB15a}.

\noindent
%\textbf{Architecture and Optimization:} 
\textbf{Optimization:} 
The architecture of the individual DIPs is similar to that used in~\cite{UlyanovVL17}.
%They further pointed that for many tasks, adding extra non-constant noise perturbations to the original constant input noise at each iteration leads to better results.
As in the basic DIP, 
%We too 
we found that adding extra non-constant noise perturbations to the input noise adds stability in the reconstruction. We gradually increase noise perturbations with iterations. We further enrich the training set by transforming the input image $I$ and the corresponding random noise inputs of all the DIPs using 8 transformations (4 rotations by $90^\circ$ combined with 2 mirror reflections - vertical and horizontal).
%These augmentations add additional training data, while maintaining the internal patch distribution and self-similarity~\cite{Zontak2011InternalSO}.
Such an augmentation was also found to be useful in the unsupervised \emph{internal learning} of~\cite{ZSSR}.
The optimization process is done using ADAM optimizer~\cite{adam}, and takes a few minutes per image on Tesla V100 GPU. \\
%Specific architecture can be found in the project code.

\noindent
\textbf{Inherent Layer Ambiguities:} \ \
\label{sec:ambiguity}
%There are often \emph{inherent} ambiguities when separating real images, which cannot be resolved without extra information.
%There are some \emph{inherent} layer separation ambiguities which cannot be resolved without extra information.
Separating a superposition of 2 pure uncorrelated textures is relatively simple (see Fig.~\ref{fig:textureSumLoss}.a). There are no real ambiguities other than a constant global color ambiguity $c$~: {$I=(y_1 + c) + (y_2-c)$}. Similarly, pure non-overlapping textures are relatively easy to segment.
However, when a single layer contains \emph{multiple independent regions}, 
as in Fig.~\ref{fig:textureSumLoss}.b,
the separation becomes ambiguous (note the switched textures in the recovered output layers of Fig.~\ref{fig:textureSumLoss}.b). 
%Such an example is shown in Fig.~\ref{fig:texture_separation}.b. 
Unfortunately, such ambiguities exist in almost any natural indoor/outdoor image.
%, both in the case of transparency separation, as well in the case of fg/bg segmentation.

To overcome this problem, initial ``hints'' are often required to guide the Double-DIP.
% with initial association of different parts of the image statistics to train the different DIPs on. 
These hints are  provided \emph{automatically} in the form of \emph{very crude} image saliency~\cite{Goferman}. Namely, in the first few iterations, \textit{DIP}$_1$ is encouraged to train more on the salient image regions, whereas \textit{DIP}$_2$ is guided to train more on the non-salient image regions. This guidance is relaxed after a few iterations.

When more than one image is available, this ambiguity is often resolved on its own, \emph{without} requiring any initial hints. For example, in \emph{video transparency}, the superposition of 2 video layers changes from frame to frame, resulting in different mixtures. The \emph{statistics} of each layer, however, remains the same throughout the video (despite its dynamics)~\cite{SelfSim_ShechtmanIrani07,SingleVideoSR2011}.
%the dynamics of the layer. 
This means that \emph{a single DIP} suffices to represent all the frames of \emph{a single video layer}. Hence, Double-DIP can be used to separate video sequences into 2 dynamic layers, and can often do so with no initial hints. 

\begin{figure}
\centering
\includegraphics[width=1.0\columnwidth]{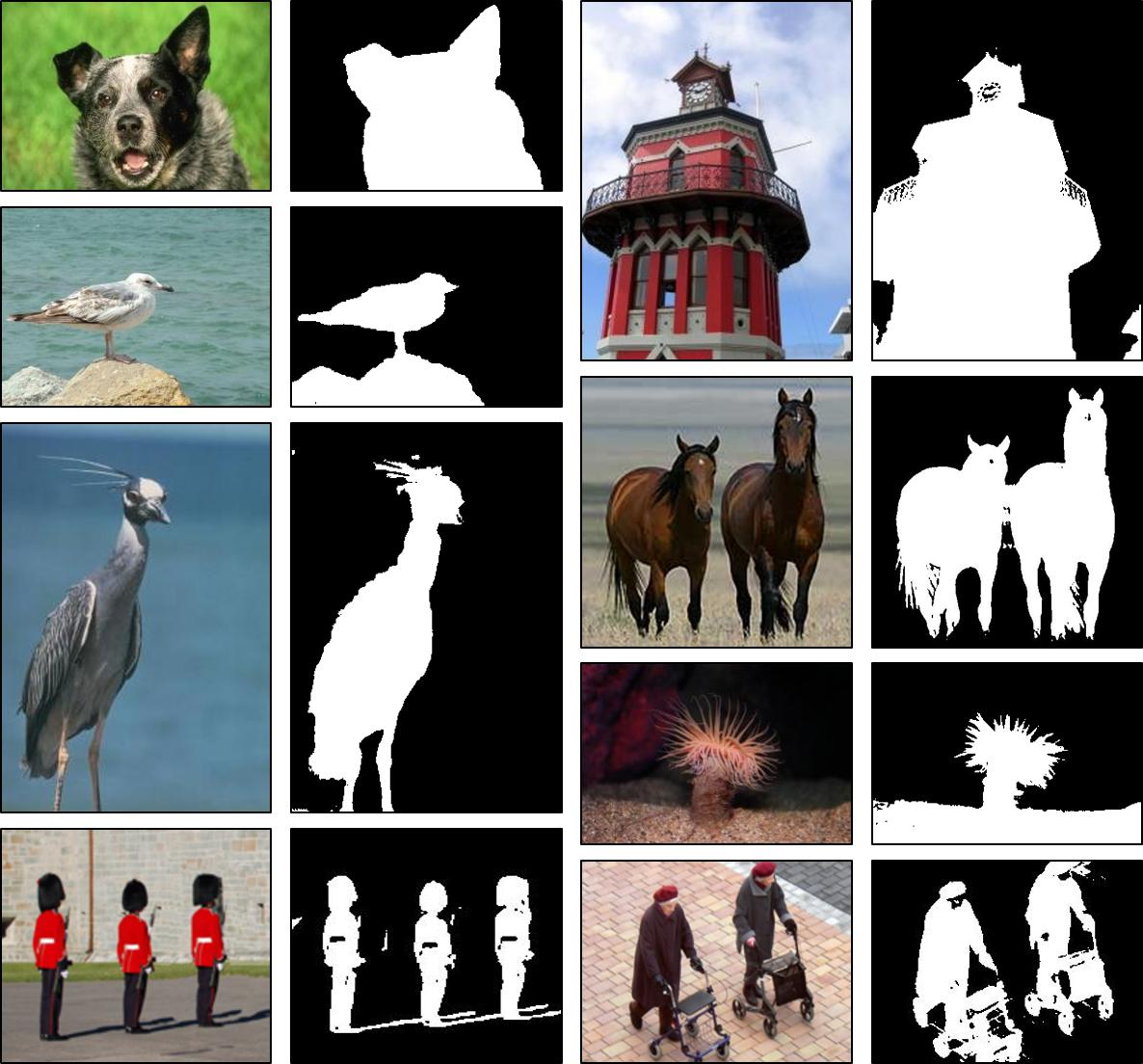}
\caption{
\textbf{Foreground/Background Image Segmentation.} 
\emph{(Please see many more results in the \href{http://www.wisdom.weizmann.ac.il/~vision/DoubleDIP/index.html}{project website})
}}
\label{fig:segmentation}
\vspace*{-0.3cm}
\end{figure}

\begin{figure}
\centering
\includegraphics[width=0.9\columnwidth]{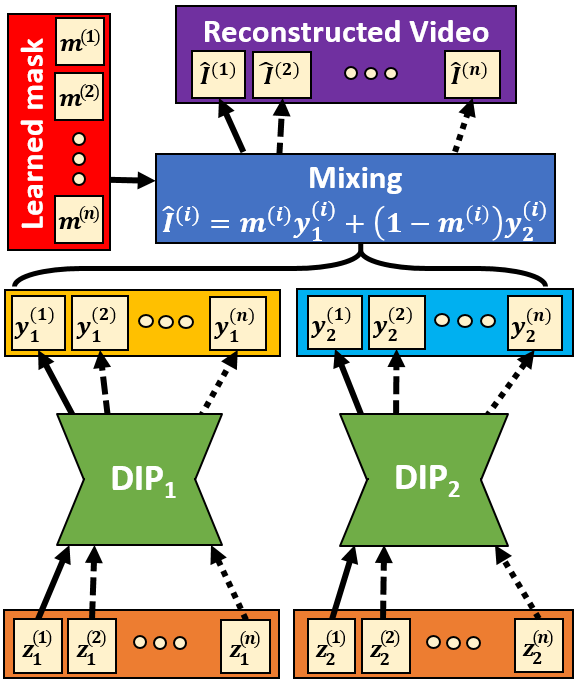}
\caption{
\textbf{Video Decomposition using Double-DIP.}  
%\emph{Full videos are attached in the supplementary material.}
}
\label{fig:video_segmentation}
\vspace*{-0.4cm}
\end{figure}

\section{Segmentation}
\label{sec:segmentation}

Fg/Bg  segmentation can be viewed as decomposing an image $I$ into a foreground layer ${y}_{1}$ and background layer ${y}_{2}$, combined by a binary mask $m(x)$ at every pixel $x$:
%for spatial coordinates $x$: 
\begin{equation}
\label{eq:mixture_seg}
I(x) = m(x){y}_{1}(x) + (1-m(x)) {y}_{2}(x)
\end{equation}
This formulation naturally fits our framework, subject to $y_1$ and $y_2$ complying to natural image priors and each being `simpler' to generate than $I$.
This requirement is verified by \cite{eccv08_bagon_boiman_irani} that defines a `good image segment' as one which can be easily composed using its own pieces, but is difficult to compose using pieces from other parts of the image. 

The Zebra image in the top row of Fig.~\ref{fig:decomposition} demonstrates the decomposition specified in Eq.~\ref{eq:mixture_seg}. It is apparent that the layers  $y_1$ and $y_2$, generated by \textit{DIP}$_1$ and \textit{DIP}$_2$,  each complies with the definition of~\cite{eccv08_bagon_boiman_irani}, thus allowing to obtain also a good segmentation mask $m$. Note that \textit{DIP}$_1$ and \textit{DIP}$_2$ automatically \emph{filled-in} the `missing' image parts in each of their output layers.

In order to encourage the learned segmentation mask $m(x)$ to be binary, we  use the following regularization loss: 
\vspace*{-0.2cm}
\begin{equation}\label{eq:l_sep_seg}    
{Loss}_{Reg}(m) = (\sum_x |m(x) - 0.5|)^{-1}
\vspace*{-0.2cm}
\end{equation}
% To stabilize our starting conditions, we initialize our learning with a crude foreground estimation based on~\cite{Goferman}. 

While Double-DIP does not capture any semantics, it is able to obtain high quality segmentation based solely on unsupervised layer separation, as shown in Fig.~\ref{fig:segmentation}. Please see many more results in the \href{http://www.wisdom.weizmann.ac.il/~vision/DoubleDIP/index.html}{project website}. Other approaches to segmentation, such as semantic-segmentation (eg., \cite{He_2017}) may outperform Double-DIP, but these are supervised and trained on many labeled examples.\\

\begin{figure}
\vspace*{-0.2cm}
\centering
\includegraphics[width=1.0\columnwidth]{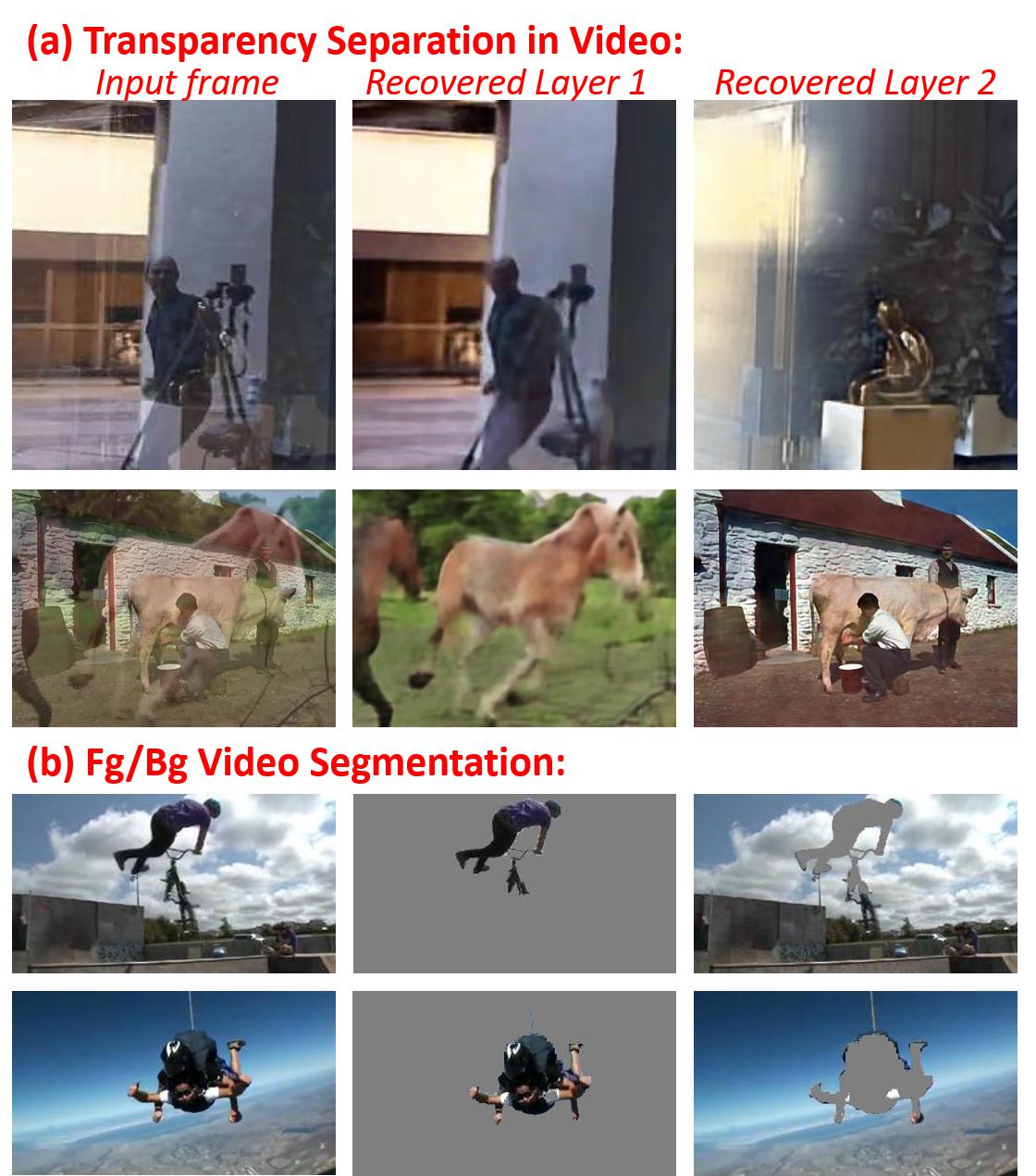}
\caption{
\textbf{Video Layer Separation via Double-DIP.}  
\emph{Double-DIP exploits the fact that all frames of a single dynamic video layer share the same patches. This promotes: (a)~video transparency separation, and (b)~Fg/Bg video segmentation. (See full videos in the \href{http://www.wisdom.weizmann.ac.il/~vision/DoubleDIP/index.html}{project website}).
%\emph{Assuming one transparent layer is static scene, while the other is dynamic video, our framework separates the layers using internal recurrence inside each layer. Full videos appear in supplementary material.
}}
\label{fig:video_separation}
\vspace*{-0.3cm}
\end{figure}

\noindent
\textbf{Video segmentation:}  
The same approach can be used for Fg/Bg video segmentation, by exploiting the fact that sequential video frames share internal patch statistics~\cite{SelfSim_ShechtmanIrani07,SingleVideoSR2011}. Video segmentation is cast as   2-layer separation 
%problem \emph{using only 2 image-based DIPs}, 
as follows:
\begin{equation}\label{eq:mixture_seg_video}
I^{(i)}(x) = m^{(i)}(x){y}^{(i)}_{1}(x) + (1-m^{(i)}(x)){y}^{(i)}_{2}(x) \quad\forall i
\end{equation}
where $i$ is the frame number. Fig.~\ref{fig:video_segmentation} depicts how \emph{a single DIP is shared by all frames} of a separated video layer:
${y}^{(1)}_{1},...,{y}^{(n)}_{1}$ are all generated by \textit{DIP}$_1$, 
${y}^{(1)}_{2},...,{y}^{(n)}_{2}$ are  generated by \textit{DIP}$_2$,
${m}^{(1)},...,{m}^{(n)}$ are all generated the mask DIP. The similarity across frames in each separated video-layer strengthens the tendency  of a single DIP to generate a consistently segmented sequence. Fig.~\ref{fig:video_separation}.b shows example frames from 2 different segmented videos (full videos can be found in the \href{http://www.wisdom.weizmann.ac.il/~vision/DoubleDIP/index.html}{project website}).

We \emph{implicitly} enforce temporal consistency in the segmentation mask, by imposing temporal consistency on the \emph{random noise} inputted to the mask DIP in successive frames:
\vspace*{-0.6cm}
\begin{equation}
{z_{m}}^{(i+1)}(x) = {z_{m}}^{(i)}(x) + \Delta{z}_{m}^{(i+1)}(x)
\vspace*{-0.2cm}
\end{equation}
where $z_m^{(i)}$ is the noise input at frame $i$ to the DIP that generates the mask. These noises change gradually from frame to frame by $\Delta{z_m^{(i)}}$ (which is a random uniform noise with variance significantly lower than that of $z_m^{(i)}$).

\begin{figure}
\vspace*{-0.2cm}
\centering
\includegraphics[width=1.0\columnwidth]{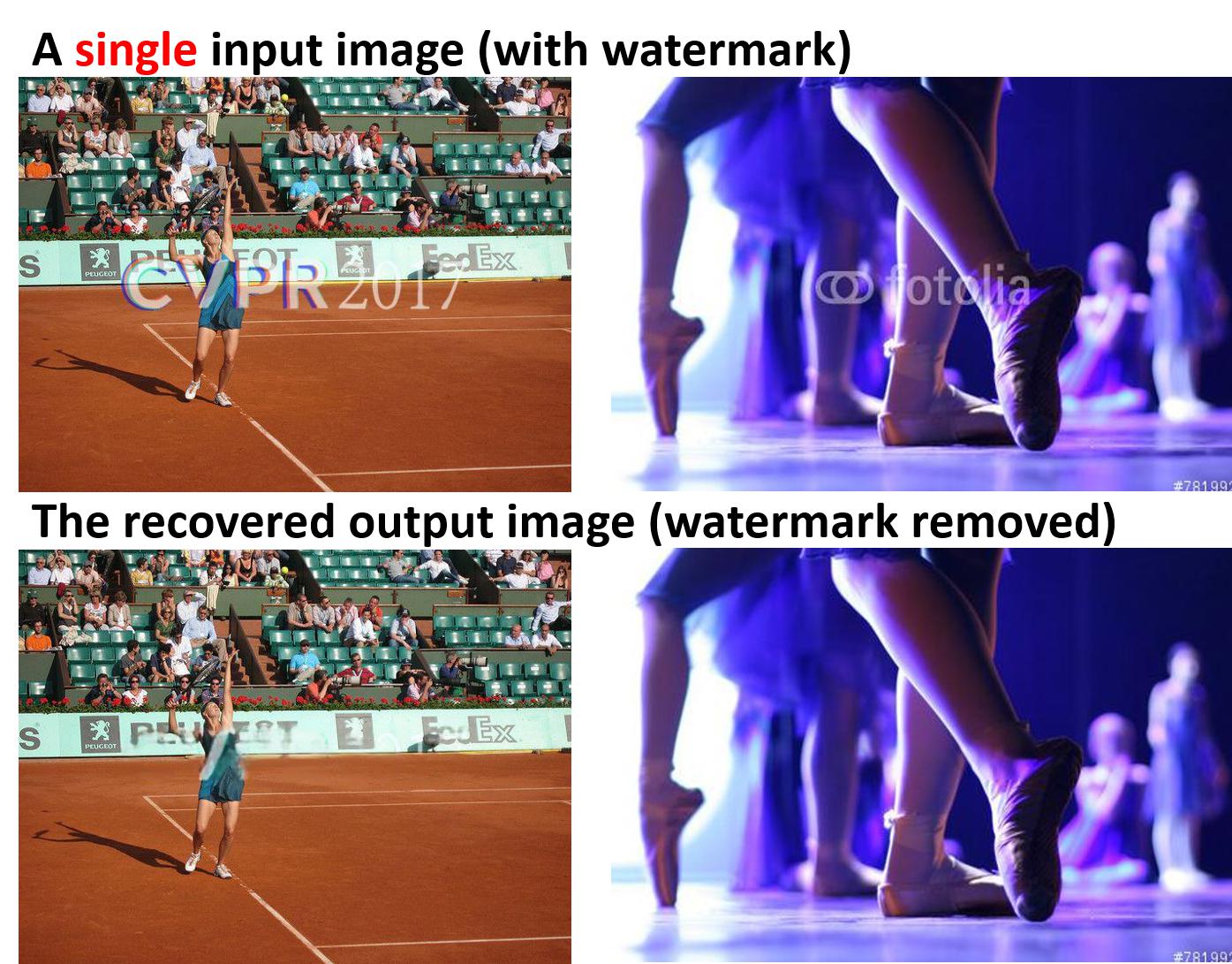}
\caption{
\textbf{Watermark removal from a single image.}
\emph{Ambiguity is resolved by a rough bounding-box around the watermark.
(Tennis image was provided in~\cite{Dekel17} as an example of an image \underline{immune} to their watermark-removal method).
}
%\emph{Given rough bounding box around the watermark in the input image, our method can be applied to remove the watermark from the image. 
%The results are presented in lower row. More results appear in supplementary material.}
}
\label{fig:watermark_one}
\vspace*{-0.3cm}
\end{figure}

\begin{figure}
\vspace*{-0.2cm}
  \centering
  \hspace*{-0.2cm} \includegraphics[width=\columnwidth]{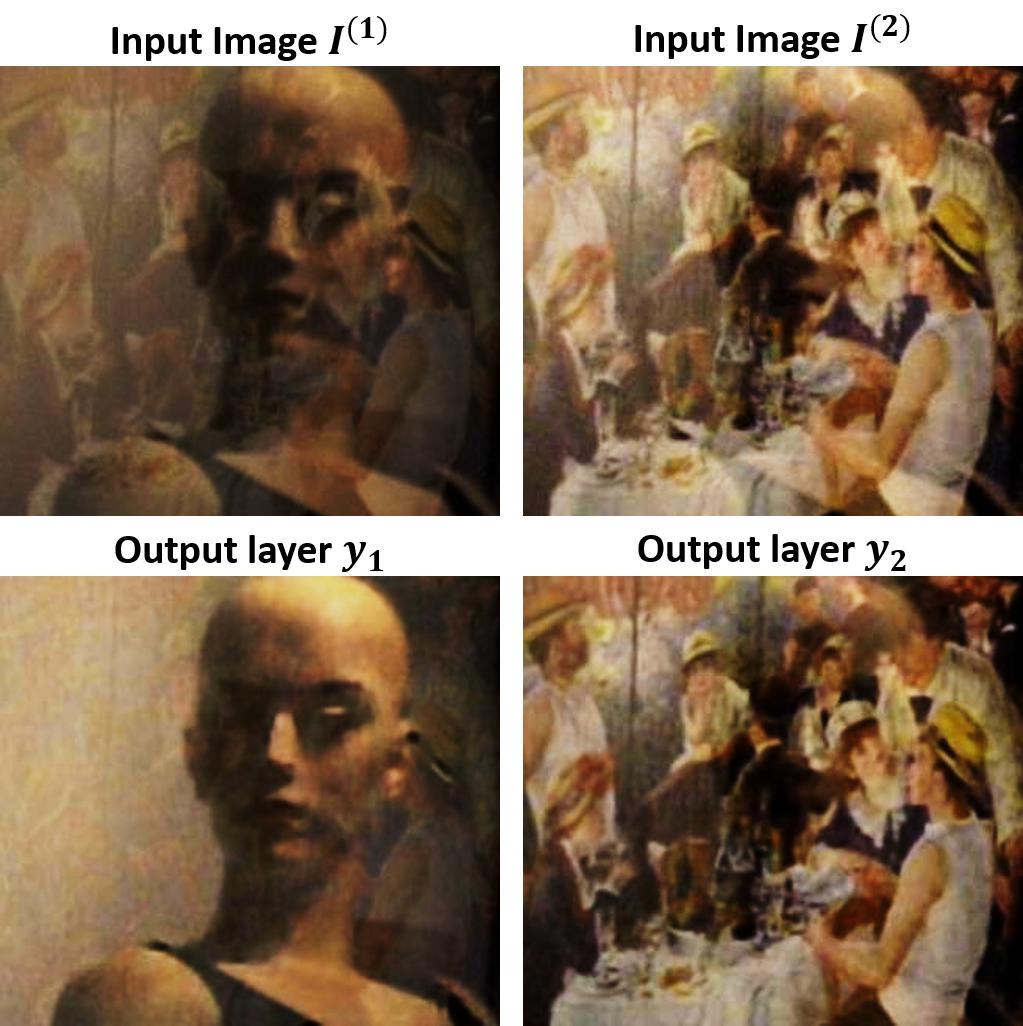}
  \vspace*{-0.1cm}
\caption{\label{fig:separation_two_images}
\textbf{Layer ambiguity is resolved when two different mixtures of the same layers are available.}
%\textbf{Two Mixtures Separation.}
%\emph{The input images are presented in the upper row. The separation by our framework are presented in the lower row.}
}
\vspace*{-0.5cm}
\end{figure}

\begin{figure*}
\vspace*{-0.2cm}
\centering
\includegraphics[width=0.9\textwidth]{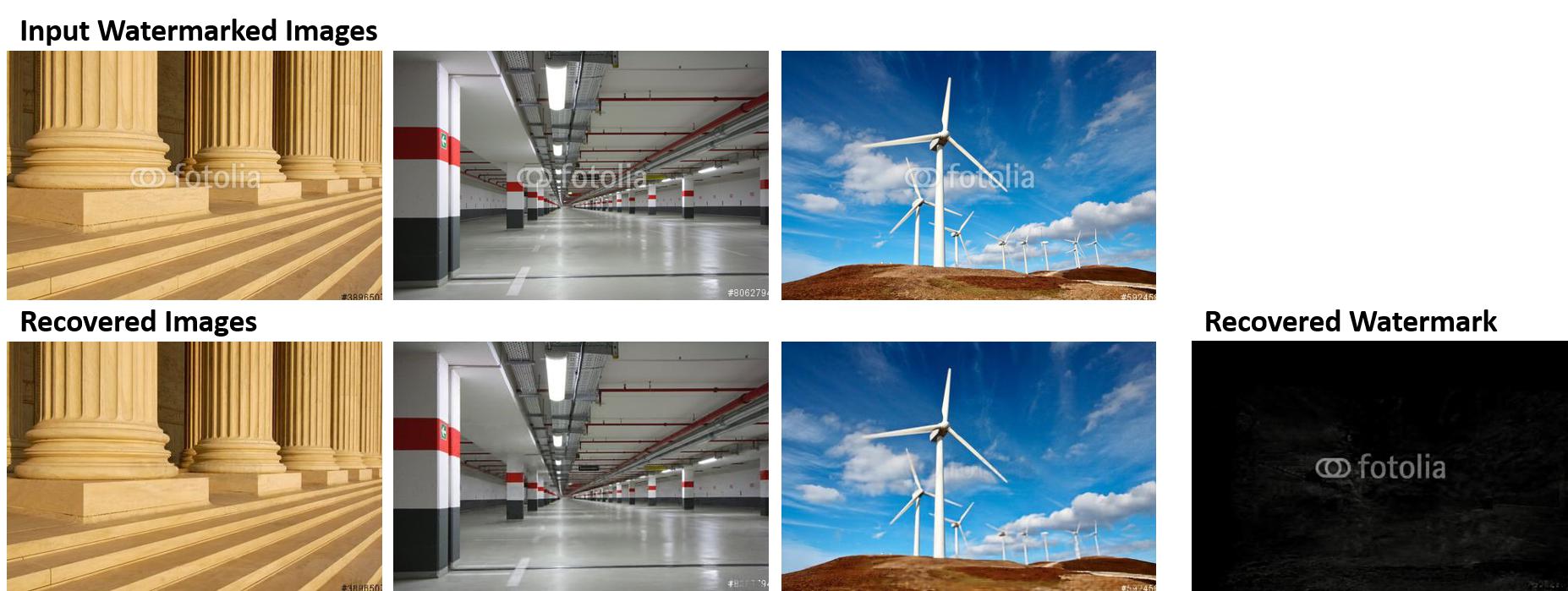}
\caption{
\textbf{Multi-image Watermark removal.} \emph{Since the 3 images share the same watermark, the layer ambiguity is resolved.}
%\emph{The framework uses 3 DIP networks to reconstruct each of the 3 input images (upper row) and another two DIP's to reconstruct the watermark and opacity. The recovered watermark is the multiplication of the reconstructed matting by the reconstructed watermark. The recovered images are the outputs of the 3 other DIP's.}
}
\label{fig:many_watermarks}
\vspace*{-0.4cm}
\end{figure*}

\section{Transparent Layers Separation}
\label{sec:tranparency}
% Many images can be viewed as combination of two or more transparent images (i.e. reflected scene through a window glass, manually added transparent watermarks, etc.). The goal of this task is to separate a given combination back to its original layers. 
% Sarel \etal \cite{Sarel} used two different transparent combinations of the same two images and minimized the correlation between the underlying layers. Levin \etal \cite{Levin2004SeparatingRF} observed that edges and corners are sparse in natural images and so the separation was generated by maximizing the sparsity of such features in each estimated layer. Li \etal \cite{Li2014SingleIL} presented separation approach based on relative smoothness of one layer.  Zhang \etal \cite{Zhang2018SingleIR} presented CNN-based method with perceptual losses and exclusion loss, closely related to gradient sparsity prior.
% Those tailor-made uses of different priors of natural images are replaced in Double-DIP with "deep image prior".  
In the case of image reflection, each pixel value in image $I(x)$ is a convex combination of a pixel from the transmission layer $y_1(x)$ and the corresponding pixel in the reflection layer $y_2(x)$. This again can be formulated as in Eq.~\ref{eq:mixture_seg}, where $m(x)$ is the reflective mask.
% \begin{equation}\label{eq:mixture_trans}
% I(x) = m(x){y}_{1}(x) + (1-m(x)) {y}_{2}(x)
% \end{equation}
%Where $I$ is the input image, $y_1$ is the transmission, $y_2$ is the reflection and 
%where $m(x)$ is the reflective mask %determining the relative convex combination at each pixel. 
In most practical cases, it is safe to assume that $m(x) \equiv m$ is a uniform mask (with an unknown constant $0<m<1$).
Double-DIP can be used to decompose an image $I$ to its transparent layers. Each layer is again constructed by a separate DIP. The constant $m$ is calculated by a third DIP. The Exclusion loss encourages minimal correlation between the recovered layers.

 Fig.~\ref{fig:textureSumLoss}.a shows  a successful separation in a \emph{simple case}, where each transparent layer has a relatively uniform patch distribution. This however does not hold in general. Because each pixel in $I$ is a mixture of 2 values, the inherent layer ambiguity  (Sec.~\ref{sec:ambiguity}) in a single transparent image is much greater than in the binary segmentation case.
 %(when no added semantic information is available).
 
Ambiguity can be resolved using \textit{external training}, as in~\cite{Zhang2018SingleIR}. However, since Double-DIP is unsupervised, 
we resolve this ambiguity when  2 different mixtures of the same layers are available. This gives rise to  coupled equations: 
%Fig.~\ref{fig:texture_separation}.a shows such separation. Due to the inherent ambiguity (\ref{sec:ambiguity}), single image separation is possible when both layers are relatively uniform. However, in many practical cases this ambiguity can be solved.
%Having two distinguish mixtures of the same layers resolves the ambiguity by giving rise to coupled equations:
\vspace*{-0.1cm}
\begin{equation}
\left\{
\begin{array}{ll}
I^{(1)}(x) = m^{(1)}{y}_{1}(x) + (1-m^{(1)}) {y}_{2}(x) \\
I^{(2)}(x) = m^{(2)}{y}_{1}(x) + (1-m^{(2)}) {y}_{2}(x)
\end{array}
\right.
\vspace*{-0.1cm}
\end{equation}
Since the layers $y_1, y_2$ are shared by both mixtures,
%$I^{(1)},I^{(2)}$, we expect a single DIP to generate an individual layer when 
one Double-DIP suffices to generate these layers
using $I^{(1)},I^{(2)}$ simultaneously. 
The different  coefficients $m^{(1)},m^{(2)}$ are generated by the same DIP using 2 random noises, $z_m^{(1)},z_m^{(2)}$
%The only difference is in the third mask-DIP, which is not shared by the two mixtures (each mixture has a different constant $m$).
%We thus employ two layer-DIPs to generate $y_1, y_2$, and two mask-DIPs to generate $m^{(1)}, m^{(2)}$. 
See such an example in Fig.~\ref{fig:separation_two_images} (real transparent images).
\\

\noindent
\textbf{Video Transparency Separation:} The case of a static reflection and dynamic transmission can be solved in a similar way. This case can be formulated as a set of equations:
\vspace*{-0.1cm}
\begin{equation}
I^{(i)}(x) = m^{(i)}{y}_{1}^{(i)}(x) + (1-m^{(i)}) {y}_{2}(x)
\vspace*{-0.1cm}
\end{equation}
where $i$ is the frame number, and $y_2$ is the static reflection (hence has no frame index $i$, but could have varying intensity over time, captured by $m^{(i)}$).
Applying Double-DIP to a separate transparent video layers is done similarly to video segmentation (Sec.~\ref{sec:segmentation}). We employ one DIP for each video layer, $y_1, y_2$ and one more DIP to generate $m^{(i)}$ (but with a modified noise input per each frame, as in the video segmentation). Fig.~\ref{fig:video_separation}.a shows examples of video separation. 
%More results and full videos can be viewed in Supplementary-Material.
For full videos see the \href{http://www.wisdom.weizmann.ac.il/~vision/DoubleDIP/index.html}{project website}.
\\

\noindent
\textbf{Watermark removal:} Watermarks are widely-used for copyright protection of photos and videos. Dekel \etal~\cite{Dekel17} presented a watermark removal algorithm, based on recurrence of the same watermark in many different images. Double-DIP is able to remove watermarks shared by very few images, often only one.

We model watermarks as a special case of image reflection,
%(Eq.~\ref{eq:mixture_seg}). 
where layers $y_1$ and $y_2$ are the clean image and the watermark, respectively. This time, however, the mask is not a constant $m$. The inherent transparent layer ambiguity is resolved by one of two practical ways: (i)~when only one watermarked image is available, the user provides a crude hint  (bounding box) around the location of the watermark; (ii)~given a few images which share the same watermark  (2-3 typically suffice), the ambiguity is resolved on its own.

When a single image and a bounding box are provided, the learned mask $m(x)$ is constrained to be zero outside the bounding box. This hint suffices for Double-DIP to perform reasonably well on this task. See examples  in Fig.~\ref{fig:watermark_one}.
In fact, the Tennis image in Fig.~\ref{fig:watermark_one} was provided by~\cite{Dekel17} as an example image immune to their watermark-removal method.

% Having multiple images containing the same watermark, each image satisfies Eq.~\ref{eq:mixture_seg}. For three such images, this can be efficiently solved: one DIP is used for the watermark $y_1$ and three more DIPs generate the three clean images $y_2^{(i)}, \quad i=1,2,3$. One last DIP is employed to resolve the opacity mask $m$, which is shared by all three images. Such examples are shown in Fig.~\ref{fig:many_watermarks}.

When multiple images contain the same watermark are available,  no bounding-box is needed. E.g., if 3 images share a watermark, we use 3 Double-DIPs, which share \textit{DIP}$_2$ to output the common watermark layer $y_2$.
Independent layers $y_1^{(i)}, i$=$1,2,3$, provide the 3 clean images. 
The opacity mask $m$, also common to the 3 images, is generated by another shared DIP.
Example is shown in Fig.~\ref{fig:many_watermarks}.

\section{Image Dehazing}
Images of outdoor scenes are often degraded by a scattering medium (e.g., haze, fog, underwater scattering). The degradation in such
images grows with scene depth.
Typically, a hazy image $I(x)$ is modeled~\cite{He2011}:
\begin{equation}\label{eq:haze}
I(x) = t(x) J(x) + (1-t(x)) A(x)
\end{equation}
where $A(x)$ is the Airlight map ($A$-map), $J (x)$ is the \emph{haze-free} image, and $t(x)$ is the \textit{transmission} ($t$-map), which exponentially decays with scene depth.
%$t(x) = exp^{−\beta Z(x)}$, where $\beta$ is a scattering coefficient and $Z(x)$ is the depth of the scene point imaged at pixel $x$.
The goal of image {dehazing} is to recover from a hazy image $I(x)$ its underlying \emph{haze-free} image $J(x)$
(i.e., the image that would have been captured on a clear day with good visibility conditions).

\emph{We treat the dehazing problem as a layer separation problem}, where one layer is the haze-free image ($y_1(x)$=$J(x)$), the second layer is the $A$-map ($y_2(x)$=$A(x)$), and the mixing mask is the $t$-map ($m(x)$=$t(x)$).\\

\begin{figure}
\centering
\includegraphics[width=1.0\columnwidth]{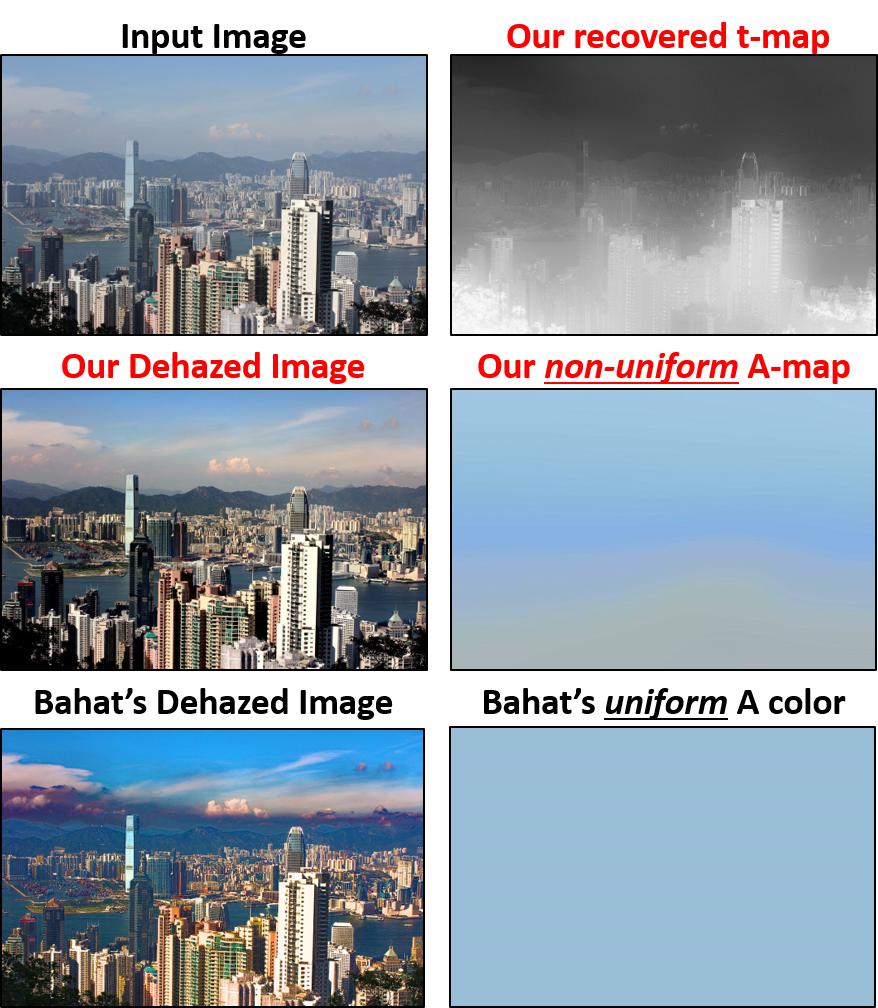}
\caption{
\textbf{Uniform vs. Non-Uniform Airlight recovery.}
\emph{The uniform airlight assumption is often violated (e.g., in dusk and dawn). Double-DIP allows to recover a non-uniform airlight map, yielding higher-quality dehazing.}
%\emph{The uniform airlight recovered by \cite{Bahat16} fails in the case of dawn or dusk, such as in this image. Our obtained non-uniform airlight is more accurate, hence the recovered dehazed image is of higher-quality.}
\label{fig:airlight}
\vspace*{-0.6cm}
}
\end{figure}

\begin{table*}
%\vspace*{-0.2cm}
\begin{center}
\begin{tabular}{|c|c|c|c|c|c|c|c|c|c|}
\hline
& He \cite{He2011} & Meng  \cite{Meng2013} & Fattal \cite{Fattal2014} & Cai \cite{CaiXJQT16} & Ancuti \cite{Ancuti} & Berman \cite{NonLocalImageDehazing} & Ren \cite{Ren-ECCV-2016} & Bahat \cite{Bahat16}& Ours \\
\hline
% \textbf{SSIM} & 0.736 & 0.754 & 0.708 & 0.667 & 0.747 & 0.750 & 0.766 & 0.726 & 0.713 \\
% \hline
\textbf{PSNR} & 16.586 & 17.444 & 15.640 & 16.208 & 16.855 & 16.610 & 19.071 & 18.640 & 18.815 \\
% \hline
% \textbf{CIEDE2000} & 20.714 & 16.847 & 19.808 & 17.295 & 16.319 & 17.112 & 14.585 & 14.123 & 14.525 \\
\hline
\end{tabular}
\end{center}
\vspace*{-0.5cm}
\caption{\small \textbf{Comparison of dehazing methods on O-Haze Dataset.}
%\emph{Although our framework is not targeted for dehazing, we achieve comparable results to main unsupervised and self-supervised dehazing algorithms.}
}
\label{table:ohaze_results}
\end{table*}

% \begin{table*}
% \vspace*{-0.2cm}
% \begin{center}
% \begin{tabular}{|c|c|c|c|c|c|c|c|c|c|}
% \hline
% & He & Meng & Fattal & Cai & Ancuti & Berman & Ren & Bahat & Ours \\
% & \etal \cite{He2011} & \etal \cite{Meng2013} & \cite{Fattal2014} & \etal \cite{CaiXJQT16} & \etal \cite{Ancuti} & \etal \cite{NonLocalImageDehazing} & \etal \cite{Ren-ECCV-2016} & \etal \cite{Bahat16} & \\
% \hline
% % \textbf{SSIM} & 0.736 & 0.754 & 0.708 & 0.667 & 0.747 & 0.750 & 0.766 & 0.726 & 0.713 \\
% % \hline
% \textbf{PSNR} & 16.586 & 17.444 & 15.640 & 16.208 & 16.855 & 16.610 & 19.071 & 18.640 & 18.815 \\
% % \hline
% % \textbf{CIEDE2000} & 20.714 & 16.847 & 19.808 & 17.295 & 16.319 & 17.112 & 14.585 & 14.123 & 14.525 \\
% \hline
% \end{tabular}
% \end{center}
% \vspace*{-0.5cm}
% \caption{\small \textbf{Comparison of dehazing methods on O-Haze Dataset.}
% %\emph{Although our framework is not targeted for dehazing, we achieve comparable results to main unsupervised and self-supervised dehazing algorithms.}
% }
% \label{table:ohaze_results}
% \end{table*}
\begin{figure*}
\vspace{-0.1cm}
\centering
\includegraphics[width=1.0\textwidth]{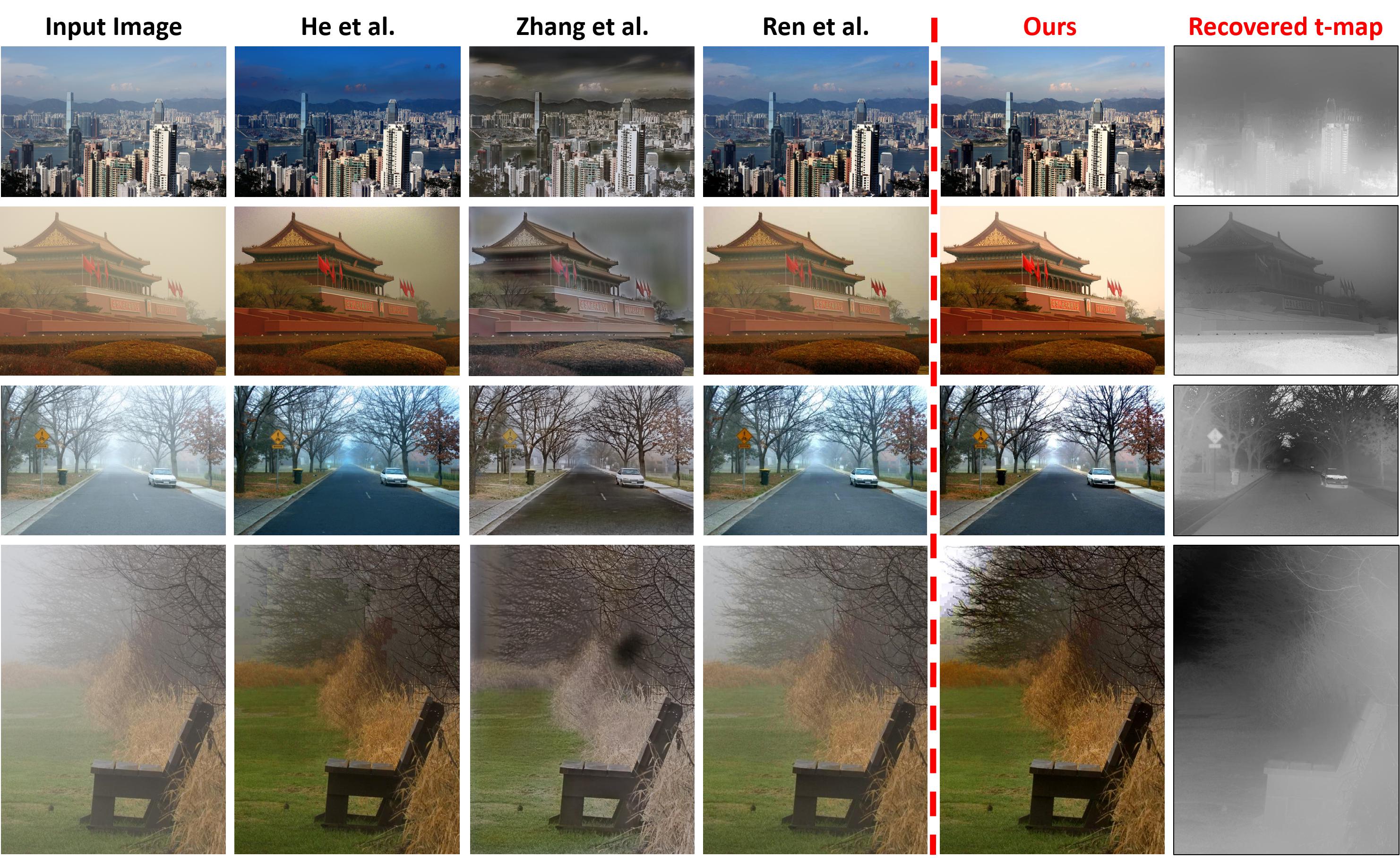}
\caption{
\textbf{Comparing Double-DIP's dehazing  to \emph{specialized} dehazing methods.} 
\emph{(many more results in the \href{http://www.wisdom.weizmann.ac.il/~vision/DoubleDIP/index.html}{project page})}
%\textbf{Visual Comparison of our framework to other dehazing methods.}
%\emph{Our framework outperforms many classic dehazing techniques such as He \etal \cite{He2011}. Note that when the methods applied to images out of O-Haze dataset \cite{O-HAZE_2018}, the algorithm proposed by Zhang \etal \cite{Zhang_2018} fails to dehaze images due to dataset overfitting.}
}
\label{fig:dehazing}
\vspace{-0.5cm}
\end{figure*}

\noindent
\textbf{Handling non-uniform airlight:} \ \ Most single-image \emph{blind} dehazing methods (e.g.~\cite{He2011,Meng2013,Fattal2014,Bahat16,NonLocalImageDehazing})  assume a \emph{uniform} airlight color $\cal{A}$ for the entire image (i.e., $A(x) \equiv \cal{A}$).
This is true also for deep network based dehazing methods~\cite{CaiXJQT16,Ren-ECCV-2016}, which train on \emph{synthesized} datasets of hazy/non-hazy image pairs.
The uniform airlight assumption, however, is only an approximation. It tends to break, e.g. in outdoor images captured at dawn or dusk,
when the sun is positioned close to the horizon.
%In such times the angle between the sun ray hitting a particle and the ray to the camera varies significantly across different image regions.
The airlight color is affected by the non-isotropic scattering of sun rays by haze particles, which causes the airlight color to vary across the image.
When the uniform airlight assumption holds, estimating a single uniform airlight color $\cal{A}$ from the hazy $I$ is relatively easy  (e.g., using the dark channel prior of~\cite{He2011}, or the patch-based prior of~\cite{Bahat16}), and the challenging part remains the $t$-map estimation.
%While it is relatively easy to estimate a single uniform airlight color $\cal{A}$ from the hazy (e.g., using the dark channel prior of~\cite{He2011}, or the patch-based prior of~\cite{Bahat16}),
However, when  the uniform airlight assumption breaks, this produces dehazing results with distorted colors. Estimating a \emph{varying} airlight, on the other hand, is a very challenging and ill-posed problem. 
Our Double-DIP framework allows simultaneous estimation of a \emph{varying} airlight-map and varying $t$-map, by treating the $A$-map as another layer, and the $t$-map as a mask. This results in higher-quality image dehazing. The effect of estimating a \textit{uniform} vs. \textit{varying} airlight is exemplified in Fig.~\ref{fig:airlight}.

%\cite{He2011} introduced dark channel prior for estimation of the transmission map and the airlight. This approach was also used by \cite{Meng2013}. Others apply different patch-based priors to reconstruct the transmission map \cite{Fattal2014, Bahat16}.
%More recent deep networks approaches use synthesized datasets of hazy and non-hazy patches \cite{CaiXJQT16} or full images \cite{Ren-ECCV-2016} for the training phase.

%Our framework applies deep networks approach, without any external or synthesized data. The model for hazy image presented above can be viewed as composition of two layers - the transmission layer and the haze-free image layer. Those layers are combined via pixel-wise multiplications and additions to reconstruct the input hazy image. ``Double-DIP'' framework can be applied to decompose back the image to those layers. This decomposition maximizes patch recurrence in each layer, as shown in \cite{Bahat16}. One DIP, $f_1$, is used to produce the clear 3-channel image $J$ and another one, $f_2$, is used to produce the airlight and is enforced to be close to some precomputed value. For experiments, we precomputed it using the method proposed in \cite{Bahat16}. The $t$-map is an output of third network $f_t$.

%${Loss}_{Reconst}= \lVert I - \hat{I} \rVert_2$, \ 
%${Loss}_{Reconst}$  in  dehazing is the $l_2$ norm between the input $I$ and the mixed output $\hat{I}$. 
In dehazing, ${Loss}_{Reg}$ 
%on $m(x)$=$t(x)$ (the $t$-map) 
forces the mask $t(x)$ to be smooth (by minimizing the norm of its Laplacian). 
%${Loss}_{Excl}$ is omitted.
%As observed in~\cite{Bahat16}, 
The  internal self-similarity of patches in a \emph{hazy image} $I(x)$ is  weaker than in its underlying \emph{haze-free} image $J(x)$~\cite{Bahat16}. This drives the first DIP to converge to a haze-free image. 
The $A$-map, however, is not a typical natural image. While it satisfies the strong internal self-similarity requirement, it tends to be much smoother than a natural image, and should not deviate much from a global airlight color. Hence, we  apply an extra regularization loss on the airlight layer: ${\lVert A(x) - \cal{A} \rVert}_2$, where $\cal{A}$ is a single initial airlight color estimated from the hazy image $I$ using one of the standard methods (we used the method of~\cite{Bahat16}). Although the deviations from the initial airlight $\cal{A}$ are subtle, they are quite crucial to the quality of the recovered haze-free image (see  Fig.~\ref{fig:airlight}).

We evaluated our framework on the O-HAZE dataset~\cite{O-HAZE_2018} and compared it to unsupervised and self-supervised dehazing algorithms. Results are presented in Table~\ref{table:ohaze_results}.
Numerically, on this dataset, we ranked second of all dehazing methods. However, visually, on images outside this dataset, our results seem to surpass all dehazing methods (see the \href{http://www.wisdom.weizmann.ac.il/~vision/DoubleDIP/index.html}{project website}).
We further wanted to compare to the winning methods of NTIRE'2018 Dehazing Challenge~\cite{ancuti2018ntire}, but only one of them had code available~\cite{Zhang_2018}. Our experiments show that while these methods obtained state-of-the-art results on the tiny test-set of the challenge (5 test images only!), they seem to severely overfit the challenge training-set. In fact, they perform very poorly on any hazy image outside this dataset (see Fig.~\ref{fig:dehazing} and the \href{http://www.wisdom.weizmann.ac.il/~vision/DoubleDIP/index.html}{project website}). A visual comparison to many more methods and on many more images is found in the \href{http://www.wisdom.weizmann.ac.il/~vision/DoubleDIP/index.html}{project website}. \\
%A visual comparison to several methods is shown in Fig.~\ref{fig:dehazing}. Many more visual examples and comparisons to additional methods can be found in the \textbf{Supplementary-Material.}
%------------------------------------------------------------------------- 

\noindent\textbf{6. CONCLUSION} 
\vspace*{0.2cm}

%\noindent
``Double-DIP'' is a unified framework for unsupervised layer decomposition, applicable for a wide variety of tasks. It needs no training examples other than the input image/video.
Although general-purpose, in some tasks  (e.g., dehazing) it achieves results comparable or even better than leading methods in the field. 
We believe that augmenting
%combining the power of 
Double-DIP with semantic/perceptual cues, may  lead to advancements also in \textit{semantic} segmentation and in other high-level tasks. This is part of our future work.
% We believe that  combining the power of 
% Double-DIP with semantic/perceptual cues, may lead to new capabilities also in high-level vision tasks. This is part of our future work.

%We believe that adding semantic/perceptual power into Double-DIP will improve its capabilities in other fields as well. This is part of our future work.

\clearpage
{\small
\bibliographystyle{ieee}

}
\end{document}